\pdfoutput=1

\documentclass[11pt]{article}

\usepackage[review]{ACL2023}
\nolinenumbers
\usepackage{times}
\usepackage{latexsym}
\usepackage{graphicx}
\usepackage[T1]{fontenc}

\usepackage[utf8]{inputenc}
\usepackage{multirow}

\usepackage{microtype}

\usepackage{xcolor}
\usepackage{inconsolata}
\usepackage[linesnumbered,ruled,vlined]{algorithm2e}

\SetCommentSty{mycommfont}
\usepackage{tikzsymbols}
\usepackage{pifont}
\usepackage[scriptsize]{subfigure}
\usepackage{amsmath,amsfonts,amssymb,amscd,amsthm,xspace}
\SetKwInput{KwInput}{Input}                
\SetKwInput{KwOutput}{Output}  

\makeatletter
\renewcommand{\tcp}[1]{\textcolor{black}{\hfill\texttt{#1}}\@hangfrom{\hskip\@totalleftmargin}{}}
\makeatother

\title{Controllable Discovery of Intents: Incremental Deep Clustering Using Semi-Supervised Contrastive Learning}

\author{Mrinal Rawat \\
  Uniphore\\
  \texttt{mrinal.rawat@uniphore.com} \\\And
  Hithesh Sankararaman \\
  Uniphore\\
  \texttt{hithesh.sankararaman}\\\texttt{@uniphore.com} \\\And
   Victor Barres\\
  Uniphore \\
  \texttt{victor@uniphore.com} 
   \\ }

\begin{document}
{\makeatletter\acl@finalcopytrue
  \maketitle
}
\begin{abstract}
Deriving value from a conversational AI system depends on the capacity of a user to translate the prior knowledge into a configuration. In most cases, discovering the set of relevant turn-level speaker intents is often one of the key steps.
Purely unsupervised algorithms provide a natural way to tackle discovery problems but make it difficult to incorporate constraints and only offer very limited control over the outcomes.
Previous work has shown that semi-supervised (deep) clustering techniques can allow the system to incorporate prior knowledge and constraints in the intent discovery process. However they did not address how to allow for control through human feedback.
In our Controllable Discovery of Intents (CDI) framework domain and prior knowledge are incorporated using a sequence of unsupervised contrastive learning on unlabeled data followed by fine-tuning on partially labeled data, and finally iterative refinement of clustering and representations through repeated clustering and pseudo-label fine-tuning. In addition, we draw from continual learning literature and use learning-without-forgetting to prevent catastrophic forgetting across those training stages. Finally, we show how this deep-clustering process can become part of an incremental discovery strategy with human-in-the-loop. We report results on both CLINC and BANKING datasets. CDI outperforms previous works by a significant margin: \textbf{10.26\%} and \textbf{11.72\%} respectively.


\end{abstract}


\section{Introduction}
Conversational AI encompasses human-machine interactions (e.g. voice assistants,  self-service bots, ...), speaker assistance during human-human conversations (e.g. customer support agent guidance, speaker coaching, ...), batch analysis of conversations, and many more use cases. Most of those make use of the concept of \texttt{`intent`} to conceptualize the relevant dimensions at the level of a conversational turn. Getting value out of those systems rests therefore in finding the intent set, i.e. the set of turn-level labels, that best reflects the practical needs of the business.

\begin{figure}[!t]
\includegraphics[trim={2.5cm 2cm 5cm 0},clip,scale=0.45]{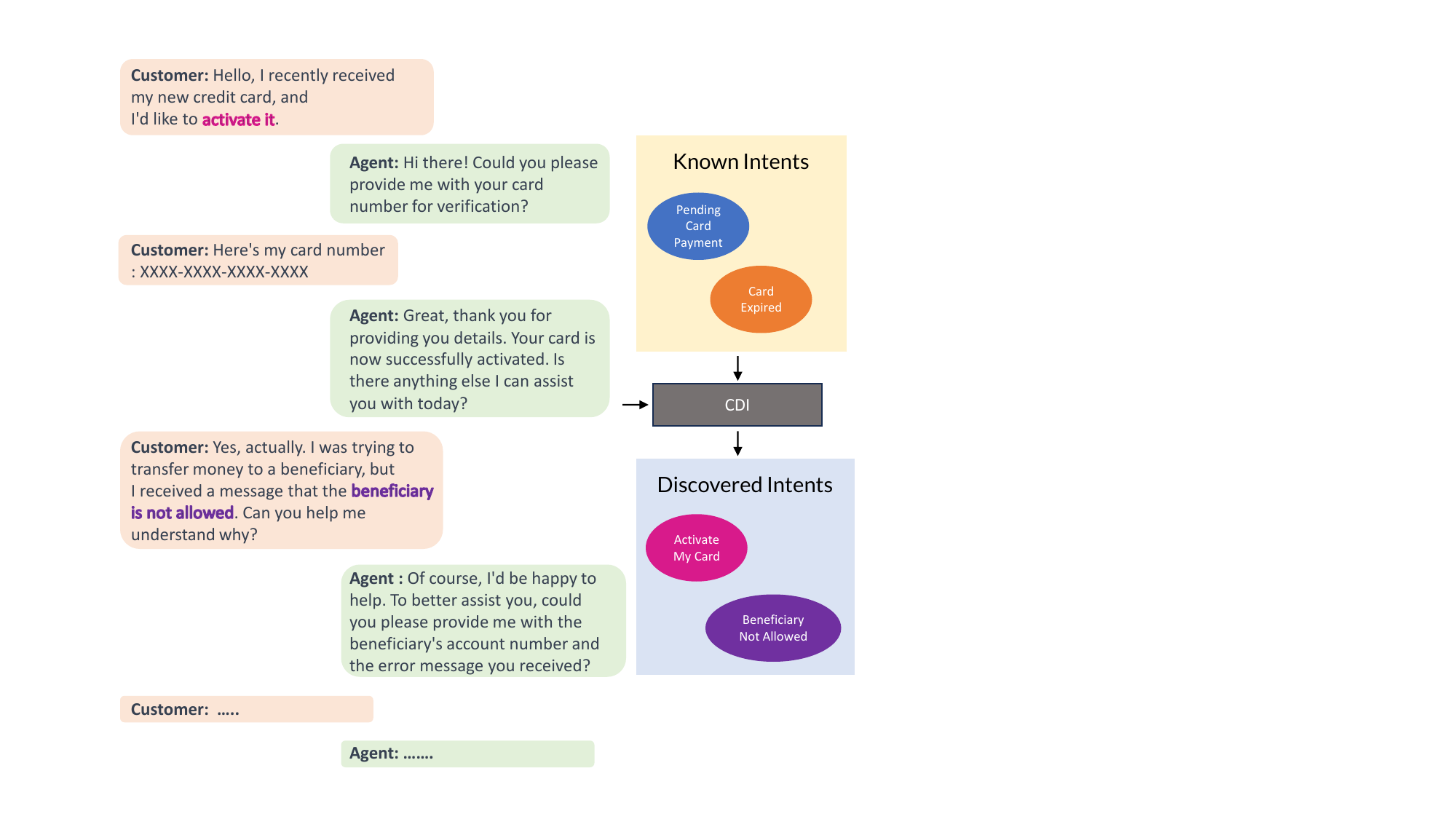}
\caption{A sample dialogue between an agent and the customer from the banking domain along with the demonstration of the intent discovery process (CDI).
}
\label{fig:arch}
\end{figure}

Businesses accumulate tacit and explicit knowledge about their processes \cite{polanyi2009tacit, nonaka2007knowledge}. But those are not often couched in ways that can be directly translated into a system's configuration. To make this possible, it is first necessary to help the user formalize their prior knowledge and processes in a way that can make them legible for a conversational AI system. In business-to-business (B2B) commercial contexts, in particular, business analysts often have to spend a large amount of time eliciting requirements from the client and compiling information prior to configuring a system. Importantly, even when formal knowledge already exists, businesses look to AI systems to help them "know what they don't already know". In the case of intents, this could take the form of helping them discover new intents to better understand their customer base, or helping them evaluate and reshape their understanding of the intent landscape (that can be sub-optimal in its current form).


Unsupervised algorithms provide a natural way to tackle such problems \cite{chatterjee-sengupta-2020-intent, benayas2023automated}.
Purely unsupervised algorithms however suffer from the fact that they lack the capacity to incorporate prior knowledge and do not offer any control over the outcome (beyond the setting of certain hyper-parameters). The objective therefore is to provide a tool that helps align an intent set with business needs. This tool should facilitate at a minimum: (1) the incorporation of domain knowledge, including the specification of required intents, and (2) the efficient intervention of an expert to guide the system toward relevant solutions.

Previous work has shown how using a combination of contrastive learning, fine-tuning, and semi-supervised learning in addition to (deep) clustering allows the system to learn to incorporate prior knowledge and constraints \cite{Zhang_Xu_Lin_Lyu_2021, shen-etal-2021-semi}. A parallel line of research has focused on using human-in-the-loop approaches to iteratively incorporate human feedback \cite{williams2015rapidly}. To our knowledge, however, no work so far has looked into combining all those elements into a single architecture.

We present a novel approach to intent discovery that satisfies the 3 requirements mentioned above.

Our contributions can be summarized as follows:
\begin{itemize}
    \item  We show how domain and prior knowledge can be incorporated using a sequence of unsupervised contrastive learning on unlabeled data followed by fine-tuning on partially labeled data, and finally iterative refinement of clustering and representations through repeated clustering and pseudo-label fine-tuning. 
    \item We show how using the learning-without-forgetting method from continual learning prevents catastrophic forgetting across those training stages, leading to improved clustering results compared to previous work. 
    \item Finally we show how this deep-clustering process can become part of an incremental discovery strategy with human-in-the-loop.
\end{itemize}

\begin{figure}[!t]
\includegraphics[trim={0.45cm 7.65cm 13cm 0},clip,scale=0.4]{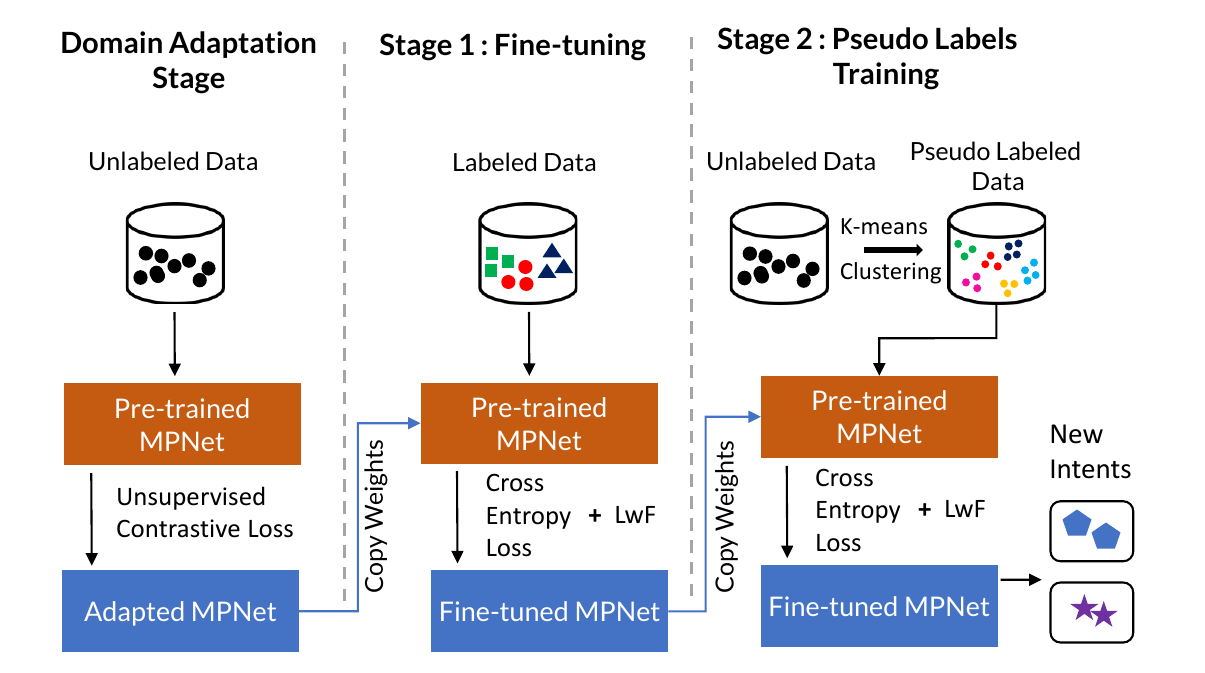}
\caption{Our proposed architecture. We begin by training the MPNet model with unsupervised contrastive loss (UCL) on the unlabeled dataset, followed by a two-stage training process along with LwF. 
}
\label{fig:arch}
\end{figure}


\section{Related Work}

Earlier works in the field of intent discovery have predominantly followed an unsupervised approach, where embeddings for the data are generated and then clustering is applied to identify new intents. However, the quality of clustering can be greatly impacted by the method used for generating input representations. Recent approaches have utilized pre-trained transformers like BERT \cite{devlin-etal-2019-bert} for generating sentence embeddings, either by extracting the [CLS] token embeddings or by mean-pooling all token embeddings. However, these methods often yield poor performance in tasks such as textual similarity and clustering, whereas sentence transformers \citep{reimers-gurevych-2019-sentence}, such as MPNet, which are trained through Siamese-based training, are more suitable for such tasks. For clustering, partition-based techniques \cite{MacQueen1967SomeMF} and density-based methods \cite{10.5555/3001460.3001507} have been proposed, but they tend to under-perform with high-dimensional data. 

Deep clustering methods overcome this problem and improve the performance significantly by jointly optimizing both input representation and clustering using deep neural networks. DEC \cite{10.5555/3045390.3045442} trains an autoencoder with reconstruction loss and iteratively optimizes the networks, while DCN \cite{10.5555/3305890.3306080} introduces a K-Means loss as a penalty term to reconstruct the clustering loss. DeepCluster \cite{10.1007/978-3-030-01264-9_9} uses the discriminative power of the convolutional neural network (CNN) and alternately performs K-Means and representation learning.

More recently, semi-supervised techniques have been widely used, such as DAC \cite{Zhang_Xu_Lin_Lyu_2021} which proposes a two-step training strategy involving supervised training using labeled samples (stage-1) followed by training samples with pseudo-labels generated by K-Means (stage-2). \citet{sahay2021semisupervised} extends the DAC work by employing a better backbone and interactive labeling. However, a limitation of such approaches is that the model may forget the learning that occurred during the first step while learning the second step, which is known as catastrophic forgetting \cite{MCCLOSKEY1989109} in literature. To address this problem \citep{DBLP:journals/corr/abs-2201-07604} retrains the model in stage-2 using the labeled dataset. In our work, we tackle this problem by incorporating the Learning without Forgetting (LwF) \cite{8107520} objective during the training process, which aims to preserve the learning that occurred during stage-1 while learning stage-2.

In addition, SCL \cite{shen-etal-2021-semi} has achieved improved results by leveraging supervised contrastive learning and a better backbone, i.e. MPNet \cite{10.5555/3495724.3497138}, in the same experimental settings. Supervised Contrastive learning \cite{NEURIPS2020_d89a66c7} involves optimizing the embedding space by pulling together the representations of samples belonging to the same class, while pushing apart the representations of dissimilar samples from other classes. However, both DAC and SCL primarily focus on labeled datasets and do not consider the use of unlabeled data.

In our work, we overcome this limitation by utilizing the unlabeled dataset and performing unsupervised contrastive learning, where positive samples are generated by passing the same sentence through the model multiple times with different dropout masks \cite{gao-etal-2021-simcse}. For positive data augmentation, other techniques, such as token shuffling and cutoff \cite{yan-etal-2021-consert}, utilizing hidden representations of BERT \cite{kim-etal-2021-self}, or back-translation \citep{Fang2020CERTCS} can also be used.

\section{Methodology}
As shown in Figure ~\ref{fig:arch} we begin with a domain adaptation step, using unsupervised contrastive learning (UCL) to adapt a sentence transformer on the unlabeled dataset. This is followed by a two-stage supervised training approach using the labeled dataset to cluster the unlabelled data and identify new intents. To ensure that the model can continuously learn and adapt to new data, we implement the learning without forgetting technique \cite{8107520}. This allows the model to incorporate new information while preserving previously learned knowledge. Further, we study the impact of enabling the incremental discovery of novel intents by incorporating human feedback in an efficient way.

\subsection{Domain Adaptation}
\subsubsection{Unsupervised Contrastive Learning (UCL)}
In Figure ~\ref{fig:arch}, we illustrate the first step of our approach: domain adaptation using unsupervised contrastive learning on the unlabeled dataset. Since in the case of unlabeled data, positive pairs are not readily available, we use the technique proposed in SimCSE \cite{gao-etal-2021-simcse}. For every input sentence $x_i$, we generate a positive pair $x_i^{+}$ by feeding the same input twice to the encoder with different dropout masks $z_i$, $z_i^{'}$. We note the embeddings $h_i^{z_i}$ and $h_i^{z_i^{'}}$.  The remaining sentences serve as negative instances. The learning objective is described below:
\begin{equation}
    \mathcal{L}_{ucl} =  -\sum_{i \in N} log \frac{e^{sim(h_i^{z_i}, h_i^{z_i^{'}}) / \tau}}{\sum_{j \neq i}^{N} e^{sim(h_i^{z_i}, h_j^{z_j^{'}}) / \tau}}
\end{equation}

for $N$ sentences mini-batch where $\tau$ is the temperature hyper parameter, and $sim(h_1, h_2)$ is the cosine similarity.

\subsubsection{Sentence Transformer}
We use a sentence transformer version of MPNet as our backbone model ('paraphrase-mpnet-base-v2'). The masked and permuted language modeling approach used to train MPNet has been shown to result in better language understanding capabilities \cite{10.5555/3454287.3454804}. The sentence transformer version is trained using the Siamese network approach pioneered by sentence BERT \cite{reimers-gurevych-2019-sentence}. 
Sentence embeddings are generated by first applying mean-pooling to the token embeddings extracted from the last hidden layer.





\subsection{Stage1: Fine-tuning}
In this first stage, we utilize a limited labeled dataset to fine-tune the model. This step allows the model to integrate the constraints and task-relevant dimensions implicitly revealed by the annotations.  This step is similar to the DAC \cite{Zhang_Xu_Lin_Lyu_2021}, with the exception that we replace the BERT backbone with the MPNet and incorporate the learning without forgetting (LwF) approach, as explained in the next subsection. We train the model using cross-entropy loss $\mathcal{L}_{ce}$ 

\begin{equation}
    \mathcal{L}_{ce} = - \frac{1}{N} \sum_{i \in N} log \frac{e^{w_{y_i}{h_i}}}{\sum_{k=1}^{K}e^{w_{y_k}h_i}}
\end{equation}

where $K$ is the number of known intents, $w$ is the classifier weights, $h_i$ is the final encoded representation and $Y (y_1, y_2,..., y_N)$ are the labels.

Once the training is complete, we remove the classifier layer and utilize the rest of the network as a feature extractor to generate sentence embeddings.

\begin{figure*}[!t]
\includegraphics[trim={2cm 5.8cm 8cm 2.5cm},clip,scale=0.7]{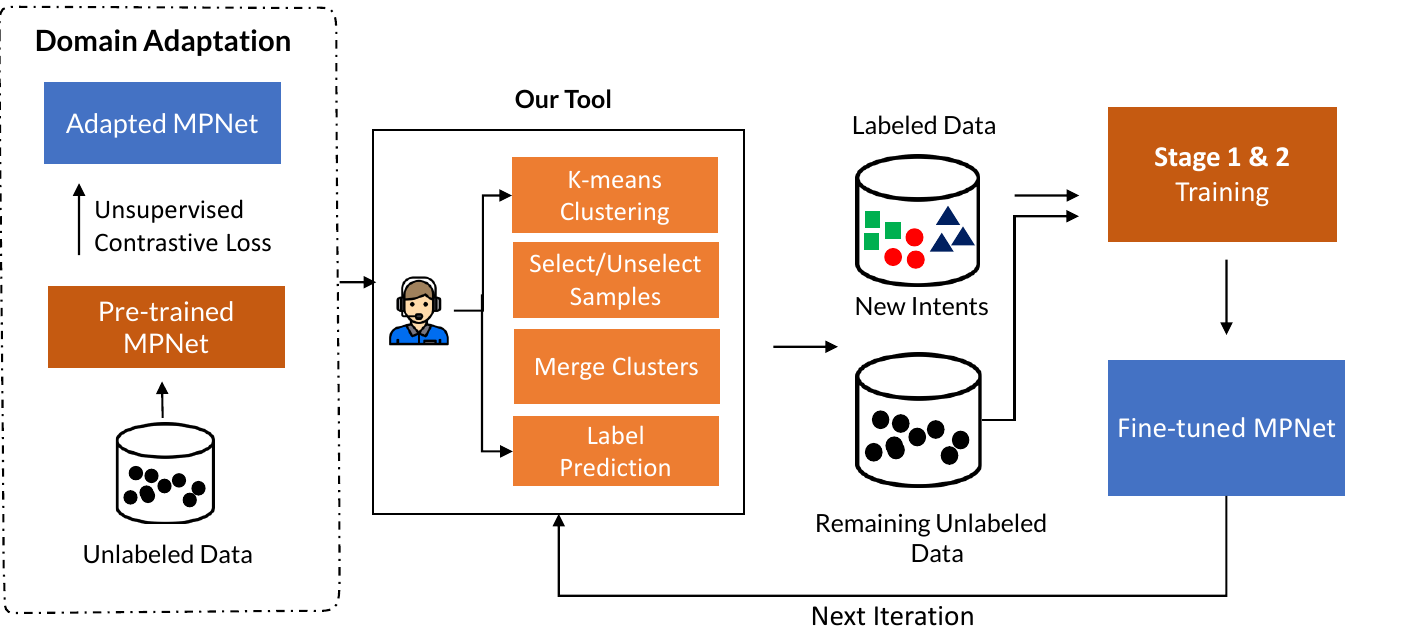}
\caption{Our proposed architecture for incremental intent discovery via human-in-the-loop involves utilizing a pre-trained model on a labeled dataset for generating representations using unsupervised contrastive learning (UCL). Then we perform K-means, and the user is presented with the clusters for input. The user can provide labeled samples and newly discovered intents by selecting or deselecting samples. Stage-1 and stage-2 training are performed using the labeled and unlabeled datasets along with LwF, and the process is iterated until no new intents are discovered.}
\label{fig:arch1}
\end{figure*}

\subsubsection{Learning without Forgetting (LwF)}
As the model learns from the labeled data, we want to ensure that it does not forget what has been learned during the domain adaptation phase: discovery of relevant new intent requires the information carried by both the domain and the labeled data to be integrated prior to clustering. 
The threat of catastrophic forgetting is a well-known threat for transfer learning approaches \cite{MCCLOSKEY1989109}. To address this problem, Learning without Forgetting (LwF) \cite{8107520} was proposed which aims to preserve the previously learned knowledge while learning new tasks. It is inspired by KL-divergence which imposes an additional constraint that the parameters of the network while learning a new task and the parameters of the old network do not shift significantly. For our work, we adopt the LwF technique and use the following objective:

\begin{equation}
    \mathcal{L}_{LwF} = - \frac{1}{N} \sum_{i=1}^{N} f(h^{'}_{i}) . log f(h_{i})
\end{equation}

\begin{align}
    f(h^{'}_{i}) = \frac{e^{w_{y_i}{h^{'}_{i}}}}{\sum_{k=1}^{K}e^{w_{y_k}h^{'}_{i}}}
& ,
     f(h_{i}) = \frac{e^{w_{y_i}{h_{i}}}}{\sum_{k=1}^{K}e^{w_{y_k}h_{i}}}
\end{align}

where $K$ is the number of known intents i.e classes, $w$ is the classifier weights, $h_i$ is the model output after learning and $h^{'}_i$ is the old model's output.

Finally, we combine this objective with the classification objective:

\begin{equation} \label{eq:L_sup}
    \mathcal{L}_{sup} = \mathcal{L}_{ce} + \lambda \mathcal{L}_{LwF}
\end{equation}
where $\lambda$ is the hyper-parameter

\subsection{Stage2: Deep-Clustering using Pseudo Labels Training}
We use the fine-tuned model to generate embeddings for all the turns in the dataset and perform K-means clustering. We assign pseudo-labels to each data point based on the K-means output and use these pseudo-labels for the supervised training of the model. Here also, to prevent catastrophic forgetting, we add an LwF objective to the cross-entropy loss.
Furthermore, this stage differs significantly from the first stage, where the number of labels or classes also changes which may lead to catastrophic forgetting. Hence, we incorporate the LwF objective alongside cross-entropy to mitigate this issue (see \ref{eq:L_sup}).

We repeat this clustering + pseudo-labeling training step multiple times. To handle the assignment inconsistency problem - the K-means cluster indices are randomly assigned at each iteration resulting in different labels - we follow the method proposed in DAC \cite{8107520} and employ the Hungarian algorithm \cite{Kuhn1955Hungarian} to align the centroids and obtain the consistent labeling. 

\subsection{Controllable Intent Discovery (CDI)}
\label{sec:hitl}
In this section, we introduce a novel approach for allowing the user to control the intent discovery process. Discovery is done in an incremental manner using our in-house developed interactive tool capturing human feedback. Our approach starts with an empty labeled dataset $D_L = \emptyset$, the unlabeled dataset $D_U$, and an empty set of Intents $I$. We perform unsupervised contrastive learning (UCL) on the unlabeled dataset $D_U$ by employing MPNet as the backbone, as shown in Figure ~\ref{fig:arch1}. Next, we choose the value of $K_t$ i.e. number of clusters. In practice, the value of K is unknown due to the lack of information about the corpus. There are various approaches of calculating the optimal value of K as proposed in previous works \cite{shen-etal-2021-semi, Zhang_Xu_Lin_Lyu_2021}. However, in this work, we did not investigate in detail to calculate the optimal value of K and used the technique proposed by DAC.

\textbf{Estimation of K}
We initialize $K'$ with a large number (e.g. 200 for our experiments which is approximately twice the largest value of intents in the datasets). However, in practical scenarios, a domain expert who uses our tool can set the initial value of K based on his understanding of the domain. Next, we use the fine-tuned model to extract intent features and perform K-means clustering. We hypothesize that real clusters tend to be dense even with a large $K'$, and that the size of more confident clusters is larger than some threshold $t$ \cite{Zhang_Xu_Lin_Lyu_2021}. Hence, we drop the low-confidence cluster which has a size smaller than t, where t is calculated as the expected cluster mean size $\frac{N}{K'}$ 



\subsubsection{Incremental Deep Clustering}
Our approach uses an incremental method to discover new intents and label the data simultaneously. At each iteration $t$, we use the trained model $M_{t-1}$ to extract representations of the unlabeled dataset $D_U$ and perform K-means clustering on these representations based on a chosen value of $K_t$. We want to highlight that at $t_1$ iteration, we use the model pre-trained with UCL loss. In the early iterations, the model may not have been fine-tuned sufficiently, leading to wrong cluster assignments for some samples. To mitigate this problem, we select only those samples which are closer to their cluster centroids, based on some threshold $\gamma$. Specifically, we compute the cosine similarity $s_i$ between each sample and its corresponding cluster centroid $C_i$ and select the sample if $s_i > \gamma$. We find that a threshold value between 0.7 and 0.99 works well in practice.

We then present the user with the resulting clusters along with the high-confidence samples sorted by confidence score $s_i$ per cluster. Our tool allows them to interactively select or deselect samples within each cluster. The user can also merge similar clusters. At the end of each iteration, we have a labeled dataset $D_t$, and a set of newly identified intents denoted as $I_{t} = (i_1, i_2, ... i_{K_{t}})$. We expand the labeled dataset as $D_L = D_L \cup D_t$ and intent set as $I = I \cup I_t$ respectively and use them to perform the stage 1 and stage 2 training along with the LwF loss to avoid catastrophic forgetting as described in above sections. In the next iteration, 
 if the number of identified intents $|I|$, exceeds the value of $K_t$, we expand and update $K_{t+1}$ to be equal to $|I|$. Otherwise, we keep $K_{t+1}$ the same as $K_t$ and continue this iterative process to discover new intents and label the data. We terminate this process once the value of $K_t$ stops increasing.




\section{Experimentation}

\subsection{Datasets}
We conduct our experiments on two public benchmark intent datasets and one private dataset. Table ~\ref{tab:stats} shows the dataset statistics.

\textbf{CLINC} is a dataset for intent classification \cite{larson-etal-2019-evaluation} that includes 22,500 queries spanning 150 intents in 10 different domains.

\textbf{BANKING} is a detailed dataset in the banking domain \cite{casanueva-etal-2020-efficient} that consists of 13,083 queries related to customer service and covers 77 distinct intents.

\textbf{TELECOM Dataset} is our private dataset which comprises of manually annotated transcripts of human-human spoken telephone conversations from the telecom customer support domain. Transcripts were generated by our in-house Kaldi-based ASR system consisting of several turns between agent and customer. In total, 1513 transcripts were collected, and for each one, our annotators identified the turn in which the caller's intent was expressed and assigned it to one of 16 pre-defined classes. However, this work only considers the intent turns as the input.

\begin{table}[h]
\centering
\resizebox{1\columnwidth}{!}{
\begin{tabular}{lllll}
\hline
Dataset & \# Classes & \# Train & \# Val & \# Test \\ \hline
CLINC   & 150        & 18000    & 2250          & 2250    \\
BANKING & 77         & 9003     & 1000          & 3080    \\
TELECOM & 16         & 1013     & 250           & 250    \\ \hline
\end{tabular}
}
\caption{Statistics of CLINC, BANKING, and TELECOM Dataset describing the number of instances used in train, validation, and test set respectively along with the number of classes.}
\label{tab:stats}
\end{table} 

\subsection{Baselines}

In our work, we conducted a direct comparison between our proposed approach and two other existing methods, namely Deep Aligned Cluster (DAC) \cite{Zhang_Xu_Lin_Lyu_2021} and Supervised Contrastive Learning (SCL) \cite{shen-etal-2021-semi}. DAC utilizes a pre-training strategy on a BERT-based backbone with limited known intent data, followed by training on pseudo-labeled data generated through a clustering algorithm. In contrast, SCL uses MPNet as the backbone and trains it on limited known intent data using a Supervised Contrastive loss \cite{NEURIPS2020_d89a66c7}. To evaluate the performance of these methods, we ran experiments and reported the results by running their code if it was available, and if not, we implemented their methods based on the description provided in their papers.

\subsection{Evaluation Metrics}
\label{sec:eval_metrics}
Following established practices in the field, for each experiment, we report the normalized mutual information (NMI), adjusted rand index (ARI), and accuracy (ACC).

\begin{algorithm}[th]
\DontPrintSemicolon
  \KwInput{Unlabeled Dataset $D_U$, Model trained using UCL $M$, True\_Intents $Y$}

\SetKwFunction{SearchF}{Evaluation}
  \SetKwProg{Fn}{Function}{:}{}

  $I \gets \emptyset$;  -> Intents \\
  $t \gets 1$; \\
$D_L \gets \emptyset$; -> Labeled Dataset\\
$K_t \gets PREDICT\_K(M, D_U)$; \\
  \While{$|I| \neq |Y|$}{
        $C_t \gets GET\_CLUSTERS(K_t, M, D_U)$
        \ForEach{$c \in C_t$}{
            $S \gets FILTER\_SENTS(\gamma)$ \tcp{High confidence samples}\\
            $S \gets$ Selects the top $75\%$ samples if all have the same label, else selects the ones having the same label in the top 20 sentences. \\
            $L \gets $ Provide Labels to the sentences from $Y$ \\
            $I \gets I \cup unique(L)$ \\
            $D_L \gets D_L \cup D(S, L)$ \\
            $D_U \gets D_U - D_L$ \tcp{Remove the samples selected for Labeled Dataset} 
        }
        $M \gets$ Model after Stage-1 \& Stage-2 training
        Get Metrics on test set \\
        \If{$|I| > K_t$}{
           $K_{t+1} \gets |I|$
        }
        \Else{
            $K_{t+1} \gets K_t$
        }
        $t \gets t + 1$
    }

\caption{Pseudo-code for automatic incremental discovery evaluation}
\vspace{-0.15cm}
\label{algo}
\end{algorithm}





\begin{table*}[!th]
\centering
\resizebox{2.05\columnwidth}{!}{
\begin{tabular}{ll|lll|lll|lll}
\hline
                                          & \multirow{2}{*}{Method} & \multicolumn{3}{c}{CLINC}                        & \multicolumn{3}{c}{BANKING}                      & \multicolumn{3}{c}{TELECOM}                     \\ \cline{3-11}
                                          &                         & ACC            & ARI            & NMI            & ACC            & ARI            & NMI            & ACC           & ARI            & NMI            \\ \hline
\multicolumn{1}{c}{\multirow{5}{*}{25\%}} & DAC (Pre-training)      & 58.8           & 44.82          & 80.09          & 43.21          & 28.82          & 62.68          & 35.6          & 17.51          & 43.22          \\
\multicolumn{1}{c}{}                      & DAC (Pseudo Training)   & 72.04          & 62.92          & 88.06          & 46.32          & 33.75          & 66.31          & 39.2          & 25.51          & 46.79          \\
\multicolumn{1}{c}{}                      & SCL                     & 74.49          & 67.77          & 90.29          & 55.19          & \textbf{44.38} & 74.68          & 41.6          & 29.99          & 46.98          \\
\multicolumn{1}{c}{}                      & CDI (Stage-1)        & 64.31          & 54.11          & 84.98          & 50.0           & 36.7           & 70.23          & 46.4          & 38.51          & 47.85          \\
\multicolumn{1}{c}{}                      & CDI (Stage-2)        & \textbf{82.27} & \textbf{75.80} & \textbf{92.97} & \textbf{57.31} & 44.02          & \textbf{75.11} & \textbf{46.8} & \textbf{38.82} & \textbf{48.98} \\ \hline
\multirow{5}{*}{50\%}                     & DAC (Pre-training)      & 70.4           & 58.48          & 85.92          & 57.76          & 44.5           & 73.13          & 44.8          & 27.35          & 49.78          \\
                                          & DAC (Pseudo Training)   & 73.8           & 64.3           & 89.08          & 56.62          & 44.41          & 73.68          & 50.0          & 35.85          & 53.94          \\
                                          & SCL                     & 77.96          & 70.53          & 91.26          & 62.63          & 50.69          & 78.5           & 56.8          & 41.9           & 56.88          \\
                                          & CDI (Stage-1)        & 77.96          & 71.37          & 91.46          & 66.27          & 53.05          & 78.97          & 60.0          & 45.87          & 58.69          \\
                                          & CDI (Stage-2)        & \textbf{86.36} & \textbf{80.97} & \textbf{94.46} & \textbf{67.56} & \textbf{56.58} & \textbf{81.06} & \textbf{62.0} & \textbf{50.44} & \textbf{60.19} \\ \hline
\multirow{5}{*}{75\%}                     & DAC (Pre-training)      & 77.64          & 69.42          & 90.29          & 65.84          & 53.55          & 78.24          & 58.8          & 40.15          & 61.2           \\
                                          & DAC (Pseudo Training)   & 84.62          & 77.07          & 93.12          & 65.71          & 53.77          & 79.52          & 54.4          & 37.94          & 60.22          \\
                                          & SCL                     & 79.27          & 72.78          & 92.15          & 63.31          & 50.69          & 78.5           & 63.6          & 48.58          & 65.27          \\
                                          & CDI (Stage-1)        & 85.87          & 79.8           & 94.12          & \textbf{75.32} & 64.02          & 84.02          & 66.8          & \textbf{55.34} & 66.26          \\
                                          & CDI (Stage-2)        & \textbf{89.87} & \textbf{85.67} & \textbf{95.86} & 75.03          & \textbf{65.55} & \textbf{85.21} & \textbf{66.8} & 54.75          & \textbf{67.86}\\\hline
\end{tabular}
}
\caption{Clustering results on CLINC, BANKING and TELECOM test dataset at known ratio of 25\%, 50\% and 75\%.}
\label{tab:part_1}
\end{table*}

\begin{figure*}[!h]
\centering
\subfigure{
\includegraphics[width=.615\columnwidth]{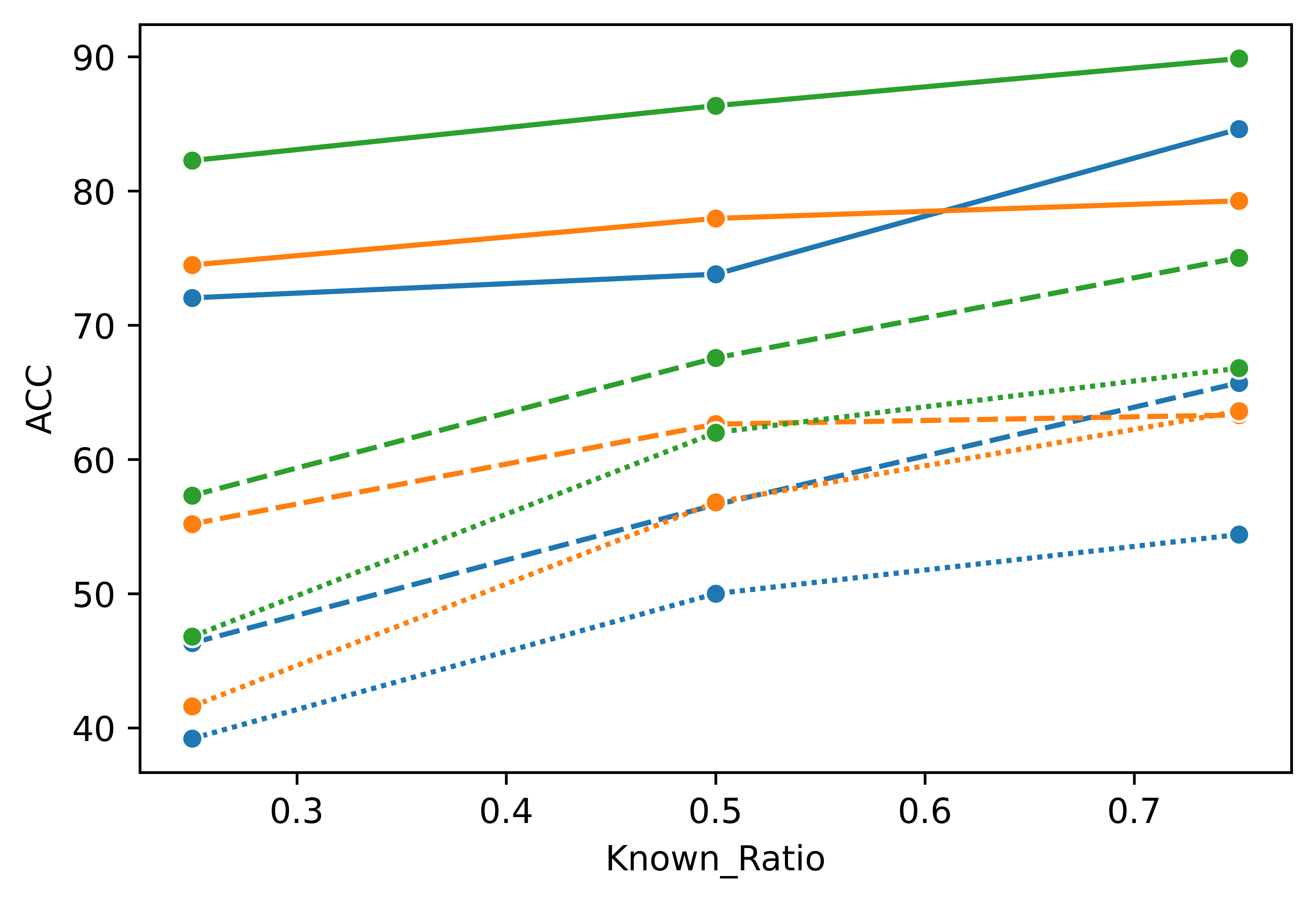}
}
\subfigure{
\includegraphics[width=.615\columnwidth]{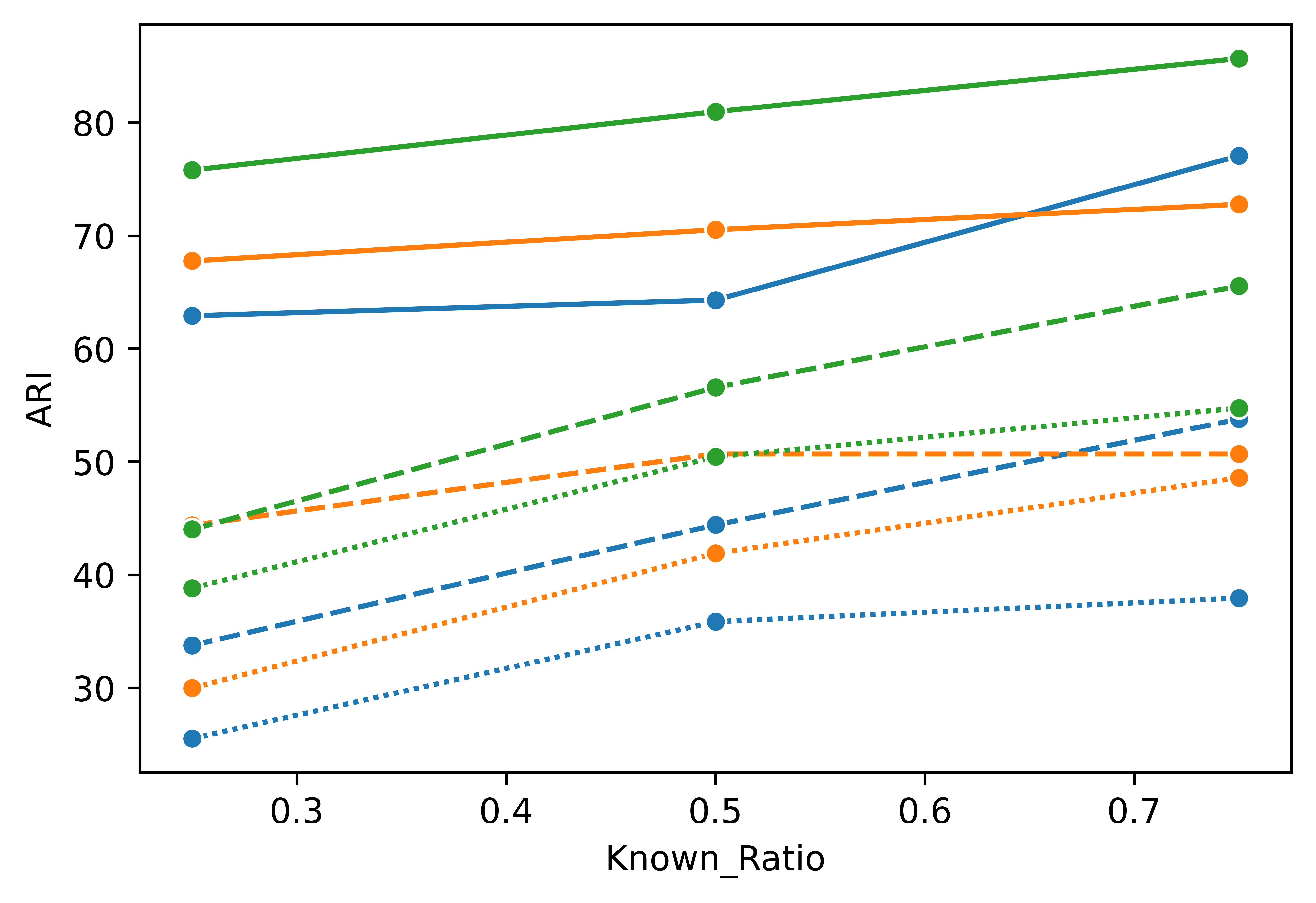}
}
\subfigure{
\includegraphics[width=.74\columnwidth]{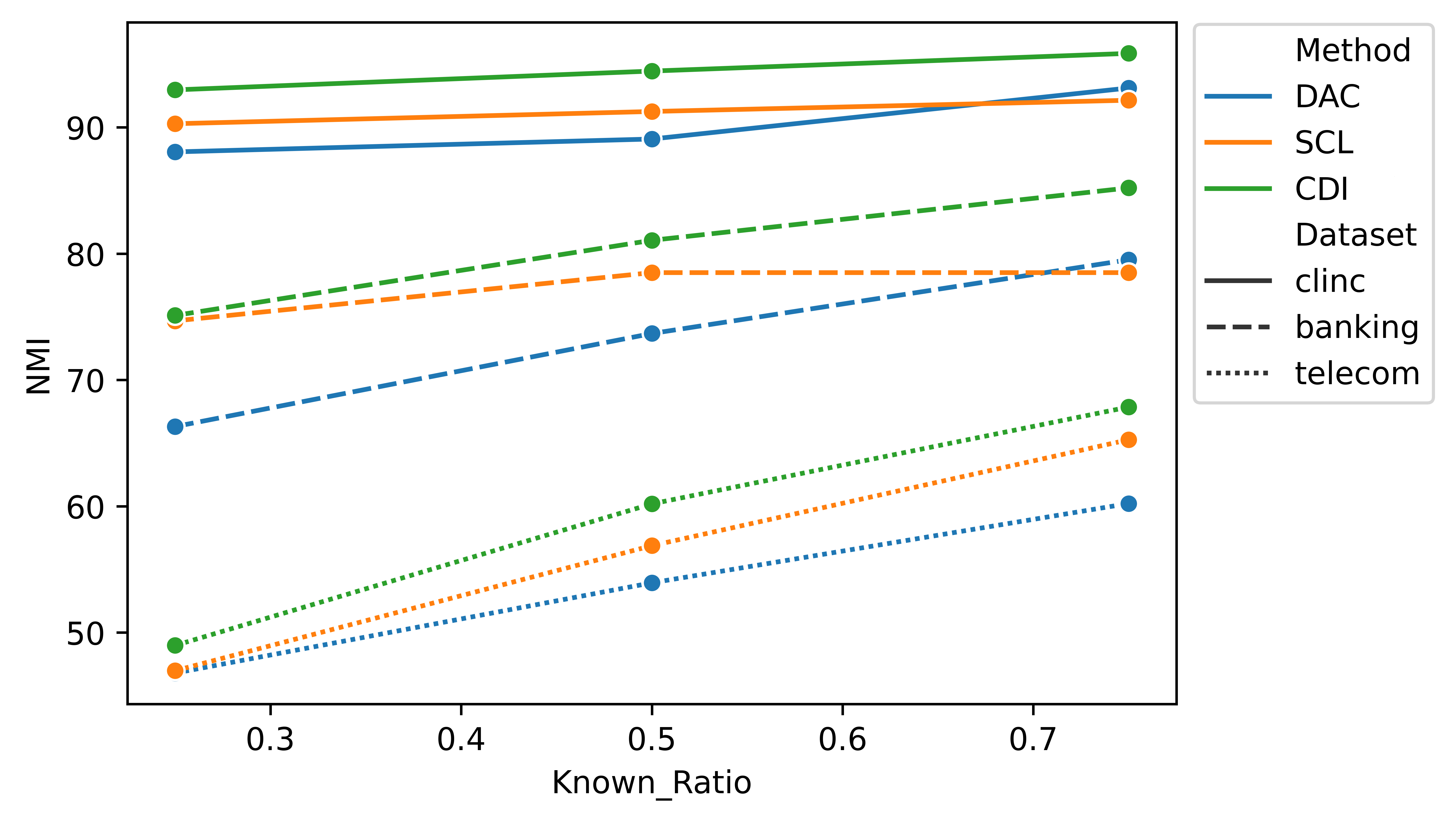}
}

\caption{Effectiveness of Known Ratio on three datasets.}
\label{fig:plot_1}
\vspace{-0.3cm}
\end{figure*}

\subsection{Evaluation Setup}
We used the same evaluation settings as defined by DAC \cite{Zhang_Xu_Lin_Lyu_2021} and CDAC \cite{lin2020discovering}. We also use the same training, validation, and test set. Our experiments were conducted with three known intent ratios of 25\%, 50\%, and 75\%. For each split, we randomly selected 10\% of samples for the CLINC and BANKING datasets, and 20\% for the TELECOM dataset, to be used as the labeled dataset. The remaining samples were treated as unlabeled data. We used the labeled dataset to train the MPnet model for multiple epochs and select the one that gives the best performance on the validation set. Our results were reported on the test set. To ensure a fair comparison, we kept the number of clusters K fixed as the ground-truth number of intents. We report the average results over five runs of experiments with different random seeds.

To evaluate the effectiveness of our incremental clustering approach, we created an automatic program that simulates a user providing input by selecting the correct samples for each cluster since we already have the ground truth labels. At the beginning of the process, we set the value of K for the first iteration based on the approach defined in Section ~\ref{sec:hitl}. Specifically, the values of $K_1$ for the CLINC, BANKING, and TELECOM datasets were estimated as 100, 50, and 10 respectively. At each iteration $t$, based on the value of K, we perform K-means clustering on the extracted representations using the pre-trained model. We then present the clusters to the automatic program along with high-confidence samples. To determine the high-confidence samples, we set a confidence threshold of $\gamma=0.7$ for the first iteration and $\gamma=0.95$ for the subsequent iterations.
To better simulate real-world scenarios, for any cluster, we select the top $75$\% samples based on the cosine distance to their cluster centroid, provided all of them have the same label. Otherwise, we only select the sentences having the same label in the top 20 sentences. This step is crucial to prevent the user from being overwhelmed with providing input in the case of heterogeneous clusters. We finally perform the stage-1 and stage-2 training and report the performance for every iteration on the test set (See Algorithm \ref{algo}). We repeat this process until we reach the ground truth value of K.


\begin{table*}[!th]
\centering
\resizebox{2.15\columnwidth}{!}{
\begin{tabular}{c|l|lllll|lllll|lllll}
\hline
\multirow{2}{*}{Iteration} & \multirow{2}{*}{Stage} & \multicolumn{5}{c}{CLINC}         & \multicolumn{5}{c}{BANKING}       & \multicolumn{5}{c}{TELECOM}             \\ \cline{3-17}
                           &                        & ACC & ARI & NMI & \%\_Labeled & K & ACC & ARI & NMI & \%\_Labeled & K & ACC  & ARI   & NMI   & \%\_Labeled & K  \\ \hline
1                          & -                      &51.6	&35.36	&77.43     &0             &100   & 46.79    &38.98     &72.36     &0             & 50 & 33.6 & 11.2  & 34.12 & 0           & 10  \\
\multirow{2}{*}{2}                          & 1                      &51.56	&43.71	&82.81         &13.97 \%             &83   &46.72	&38.22	&71.31    &16.92 \%             &44   & 40.0 & 17.93 & 33.14 & 3.4 \%      & 5  \\
                          & 2                      &52.0	&49.87	&85.8     &13.97 \%             &83   &47.44	&38.18	&72.84   &16.92 \%             &44   & 42.4 & 24.22 & 42.4  & 3.4 \%      & 5  \\
 \multirow{2}{*}{3}                         & 1                      &65.2	&57.93	&87.74             &26.73 \%     &105   &61.4	&50.89	&79.19     &24.31 \%             &60   & 51.2 & 35.17 & 51.45 & 12.03 \%    & 8  \\
                          & 2                      &63.96	&59.51	&89.43             &26.73 \%    &105  &62.66	&50.79	&78.68     &24.31 \%             &60   & 54.4 & 36.74 & 53.87 & 12.03 \%    & 8  \\
\multirow{2}{*}{4}                         & 1                  &77.11	&69.99	&91.52     &40.51 \%             &129   &68.83	&57.15	&81.62     &35.96 \%             &66   & 63.6 & 46.53 & 61.41 & 20.34 \%    & 9  \\
                          & 2                &79.73	&74.64	&93.31               &40.51 \%   &129    &67.01	&56.02	&81.64     &35.96 \%              &66   & 65.6 & 49.07 & 62.02 & 20.34 \%    & 9  \\
\multirow{2}{*}{5}                          & 1       &88.71	&84.14	&95.56     &63.75 \%             &143   &73.57	&63.21	&84.56     &47.10 \%             &70   & 67.6 & 52.68 & 64.33 & 24.22 \%    & 10 \\
                         & 2    &88.58	&83.62	&95.48     &63.75 \%             &143   &73.02	&63.08	&84.72     &47.10 \%             &70   & 71.6 & 55.79 & 67.33 & 24.22 \%    & 10 \\
\multirow{2}{*}{6}                         & 1   &92.13	&88.36	&96.72     
&82.78 \%             &147   &82.11	&71.93	&87.53     &59.12 \%             &73   & 72.4 & 59.56 & 69.75 & 41.01 \%    & 13 \\
                          & 2     &93.6	&90.18	&97.3     &82.78 \%             &147   &78.73	&68.21	&86.51     &59.12 \%             &73   & 74.0 & 58.79 & 71.17 & 41.01 \%    & 13 \\
\multirow{2}{*}{7}                          & 1      &\textbf{96.76}	&94.63	&98.36     &91.43 \%             &150   &84.45	&75.07	&89.15     &70.69 \%             &76   & 72.8 & 58.2  & 70.83 & 54.71 \%    & 14 \\
                          & 2         &96.62	&\textbf{94.78}	&\textbf{98.42}     &91.43 \%             &150   &85.29	&75.56	&89.43     &70.69 \%             &76   & 72.8 & 55.89 & 69.72 & 54.71 \%    & 14 \\
\multirow{2}{*}{8}                          & 1                      &  -   &   -  &  -   &      -       & -  &\textbf{90.39}	&\textbf{81.72}	&\textbf{91.18}     &81.78 \%             &77   & 76.4 & 60.78 & 73.97 & 60.17 \%    & 15 \\
                          & 2                      &   -  &  -   &  -   &      -       & -  &87.53	&78.71	&90.28    &81.78 \%             &77   & 75.6 & 62.56 & 72.89 & 60.17 \%    & 15 \\
\multirow{2}{*}{9}                          & 1                      & -    & -    & -    &             - & -  &   -  &     - & -     &             -& -  & 78.0 & 68.16 & 75.07 & 71.81 \%    & 16 \\
                          & 2                      &  -    &     - & -     & -             & -   & -     &      -& -    &              -& -   & \textbf{78.8} &\textbf{ 68.56} & \textbf{75.26} & 71.81 \%    & 16 \\ \hline
\end{tabular}
}
\caption{Iteration-wise clustering results for the human-in-the-loop approach on three datasets. K indicates the number of intents discovered, while \%\_Labeled indicates the percentage of labeled samples until that iteration. We stop reporting the results when K reaches the ground truth intent value.}
\label{tab:part-2}
\end{table*}



\begin{figure*}[h!]
\centering
\subfigure[Iteration 1, K=0]{
\includegraphics[width=.48\columnwidth]{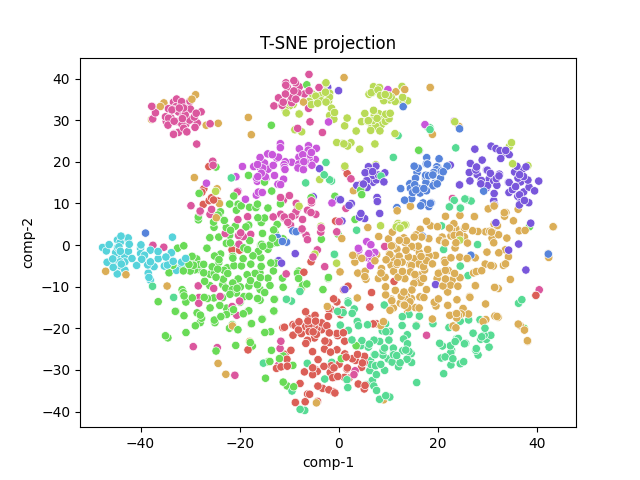}
}
\subfigure[Iteration 2, K=5]{
\includegraphics[width=.48\columnwidth]{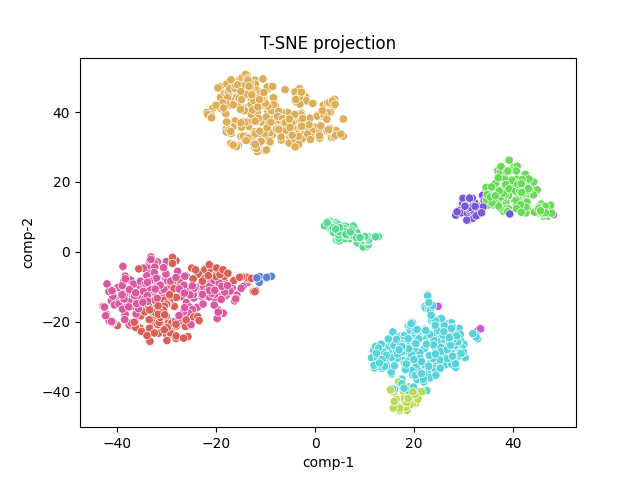}
}
\subfigure[Iteration 7, K=14]{
\includegraphics[width=.48\columnwidth]{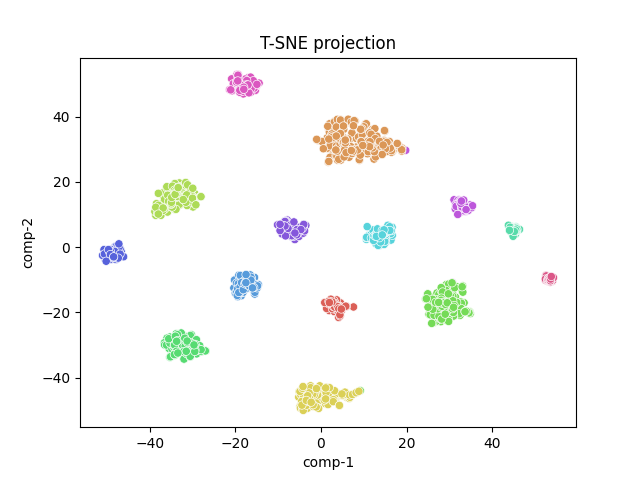}
}
\subfigure[Iteration 9, K=16]{
\includegraphics[width=.48\columnwidth]{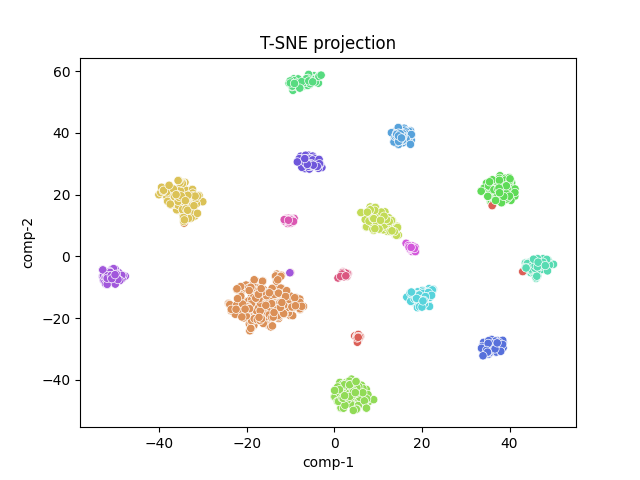}
}
\caption{TSNE plots at different iterations for TELECOM dataset.}
\label{fig:plot_tsne}
\vspace{-0.3cm}
\end{figure*}

\subsection{Training Details}
We utilized the MPNet model \cite{reimers-gurevych-2019-sentence} as the backbone for both stage-1 and stage-2 and adopted most of its hyper-parameters for the optimization. We freeze the initial 11 layers of the model and only perform learning on the subsequent layers. To improve the learning capacity of our model, we add a dense layer followed by a Tanh activation function. The dimension of the sentence representation was set to 768, while the learning rate was 5e-5, and the batch size depended on the GPU's availability. Moreover, we set $\gamma$ as 0.75 for the first iteration and 0.95 for the subsequent iterations to select high-confidence samples. To incorporate the LwF objective, we set $\lambda$ as 0.5 in both stages. All models were implemented in PyTorch using HuggingFace's transformers library \cite{DBLP:journals/corr/abs-1910-03771}.

\section{Results \& Discussion}
Our evaluation is based on the metrics specified in Section ~\ref{sec:eval_metrics} on the test set. Our findings are presented in two parts: 1) a comparison of our results with those of the previous state-of-the-art works, utilizing the same settings as proposed by them, and 2) an evaluation of our human-in-the-loop approach.

Table \ref{tab:part_1} illustrates the key findings from our part-1 experiments. Our model consistently outperforms the strongest baseline DAC by a significant margin on all three datasets. However, there are some cases, such as the BANKING dataset with a known labeled ratio of 25\%, where SCL performed better with a very small margin of \textbf{0.36\%}. Notably, on the CLINC dataset, our model achieves a good accuracy of \textbf{82.27\%} even with just 25\% known classes ratio, surpassing DAC and SCL by \textbf{10.23\%} and \textbf{7.78\%} respectively. It is worth noting that SCL generally performed better than DAC in most cases, which could be attributed to the choice of backbone as MPNet, instead of BERT, as MPNet is fine-tuned on the similarity measure.

Next, we observe that our stage-2 training demonstrates significant performance improvement as compared to stage-1 in most cases. This highlights the effectiveness of incorporating the Learning without Forgetting (LwF) objective, as stage-2 involves a different task than stage-1, and LwF prevents forgetting from occurring. Furthermore, we found that in scenarios where the labeled dataset was limited, such as the 25\% known ratio, supervised contrastive learning (SupCon) used in SCL outperformed our stage-1 in both the CLINC and BANKING datasets. This indicates that SupCon is beneficial when dealing with limited labeled data, as it enhances class separability. However, as more intents become known, the additional benefit of SupCon diminishes and our approach performs significantly better.

\subsection{Results with Human-in-the-loop}
Table \ref{tab:part-2} presents the results of our incremental intent discovery approach, which incorporates simulated human-in-the-loop feedback. As illustrated in Table \ref{tab:part-2}, we report the performance in both stages for each iteration. In the first iteration, we start with all unlabeled data and set the value of K using the method described in the previous sections. Subsequently, in each iteration, we continue to discover new intents, label data, and improve the performance metrics, including ACC, ARI, and NMI, on the test set. We terminate the process when K reaches the ground truth number of intents. Specifically, for the CLINC dataset, it took us \textbf{7} iterations to label \textbf{91.43\%} of the dataset, \textbf{8} iterations for the BANKING dataset to label \textbf{81.78\%} of the dataset, and \textbf{9} iterations for the TELECOM domain dataset to label \textbf{71.81\%} of the dataset.  Additionally, Figure \ref{fig:plot_tsne} illustrates the tsne plot for each iteration, showcasing the separation of samples class-wise and the addition of new intents in subsequent iterations on the TELECOM dataset.


\section{Conclusion \& Future Work}

In this work, we present a Controllable Discovery of Intents (CDI) framework where prior knowledge is incorporated using unsupervised contrastive learning followed by a two-stage fine-tuning strategy. We also propose a novel incremental intent discovery method that incorporates human-in-the-loop feedback, while also utilizing the learning without forgetting (LwF) objective to preserve previously learned knowledge during new iterations. Our experimental results demonstrate that our approach significantly outperforms previous works by a significant margin. In future work, we plan to extend our approach to other languages and explore its applicability to entity discovery.

\section*{Limitations}
Our work has certain limitations that should be acknowledged. Turn embeddings do not account for the larger context of the transcript in which the turn appears. In conversational datasets such as TELECOM, incorporating such contextual information can potentially improve performance. The candidate selection method would benefit from being more thoroughly investigated. Using the distance to the clusters centroids to select candidates with a high threshold may result in a reduced number of sentences selected per cluster, leading to decreased efficiency. 
The human-in-the-loop component is evaluated by simulating the user. We see efficient and standardized ways of automatically testing systems that incorporate human feedback as key to accelerate the development of such architectures. Future work will however need to focus on running real user experiments both to validate our current approach as well as to improve the automatic testing procedure.

\bibliography{custom}
\bibliographystyle{acl_natbib}

\appendix



\end{document}